\documentclass[runningheads]{llncs}
\usepackage[T1]{fontenc}
\usepackage{graphicx}
\newcommand{\shortus}{\scalebox{0.55}[1]{\_}} 
\usepackage{amssymb,amsmath}
\usepackage{algorithm}
\usepackage{algorithmic}
\usepackage{booktabs,multirow}
\usepackage{tabularx}
\usepackage{fontawesome}
\newcolumntype{Z}{>{\centering\arraybackslash}X} 
\usepackage{subfig}   
\usepackage{float}

\usepackage[skip=4pt]{caption} 

\usepackage{xcolor}

\usepackage{hyperref}
\hypersetup{
    unicode=true,
    colorlinks=true,      
    linkcolor=blue,       
    citecolor=blue,       
    urlcolor=blue,        
    breaklinks=true,      
    bookmarksnumbered=true
}

\usepackage{caption}
\begin{document}
\setlength{\textfloatsep}{5mm}  

\setlength{\abovecaptionskip}{3mm}  
\setlength{\belowcaptionskip}{3mm}  

\title{Online Signature Verification Using Augmented Path Signature and T-Mamba}
\titlerunning{Online Signature Verification Using APS and T-Mamba}
%
\author{Ruiling Li \and Danyu Yang\textsuperscript{(\faEnvelopeO)}}
\authorrunning{R. Li and D. Yang}
%
\institute{College of Mathematics and Statistics, Chongqing University, China
\email{202406021058T@stu.cqu.edu.cn, danyuyang@cqu.edu.cn}\\ }
%

\maketitle              

\begin{abstract}
Handwritten signature verification is vital for personal authentication across commercial and financial applications. Although deep learning methods are widely adopted for online signature verification (OSV), they often struggle with capturing highly discriminative features and modelling long-range dependencies. To address these issues, we propose a novel framework that integrates the augmented path signature (APS) descriptor with the T-Mamba model. The APS descriptor first applies time and basepoint augmentations, then computes sliding-window path signatures. The path signature is a non-parametric feature map from rough path theory that effectively captures geometric structures and nonlinear inter-channel interactions. Inspired by the efficacy of state space models (SSMs) in sequence modelling, our T-Mamba model employs a hybrid design combining two temporal convolutional network (TCN) blocks with a time-scanning Mamba. This design enables the model to learn both local temporal patterns and global long-range dependencies, substantially improving verification accuracy. Our framework achieves state-of-the-art EERs on three public benchmark datasets (MCYT-100, SVC-2004 Task 2, DeepSignDB), validating its effectiveness and robustness, especially when the training data is limited. Our code is publicly available at \url{https://github.com/DLRL04/OSV-using-APS-and-T-Mamba}.
\keywords{Online signature verification \and Mamba \and Temporal convolutional network \and Path signature \and Dynamic time warping.}
\end{abstract}
\section{Introduction}
In the past decades, handwritten signatures have served as one of the most widely accepted forms of personal authentication, playing a vital role in administrative, commercial, and financial contexts such as contract signing and payment authorization.
Signature verification  is typically categorized as offline and online approaches based on the data acquisition process. While offline signature verification systems analyse static images, online signature verification (OSV) systems capture dynamic trajectories using specialised digital devices such as smartphones and pressure-sensitive tablets, making them more reliable than offline approaches~\cite{plamondon2000online,impedovo2012handwritten}.
However, effectively modelling the complex and  multidimensional time series for OSV remains challenging. This difficulty arises from the substantial intra-writer variability inherent in human handwriting, as well as the presence of skilled and random forgery attacks, making OSV considerably more challenging than many other biometric modalities~\cite{sundararajan2018deep}.

Traditional parameter- or function-based feature extraction for OSV lacks explicit mechanisms to address intra-writer variability. To overcome this, we introduce the augmented path signature (APS) descriptor as a robust, discriminative representation. Specifically, raw signature trajectories are augmented with timestamps and basepoints prior to sliding-window path signature computation~\cite{chevyrev2026primer}. Grounded in rough path theory~\cite{lyons1998differential}, this principled non-parametric descriptor effectively captures local geometric structures and nonlinear inter-channel interactions, thereby significantly boosting verification performance.

With the advent of deep learning, models such as RNNs~\cite{li2019stroke,lai2017online}, CNNs~\cite{bhowal2022two,vorugunti2019online}, and Transformers~\cite{gautam2023tsosvnet,melzi2023exploring} have achieved promising results in OSV. Nevertheless, RNNs are capable of capturing temporal dependencies but struggle with long sequences and typically require a substantial amount of training data. CNNs efficiently extract local and spatial patterns but are limited by their receptive fields. While Transformers excel at modelling long-range dependencies, they  incur high computational costs and may overfit when training data is limited.
Although great advances have been made, signature verification remains an open challenge.
In light of this, we introduce T-Mamba, a hybrid model that integrates temporal convolutional networks (TCN) with time-scanning Mamba, a selective state-space model based on Mamba~\cite{gu2024mamba}. By employing TCNs with dilated convolutions, we first expand the receptive field to capture fine-grained local patterns. Then the time-scanning
Mamba processes the sequence in both forward and backward temporal directions with shared parameters, enabling globally selective integration of relevant information with linear complexity. This combination keeps T-Mamba computationally efficient and robust to the variable lengths typical of handwritten signature sequences.

Alongside representation learning, deep metric learning has emerged as a cornerstone of signature verification by optimizing feature spaces to maximize inter-writer separability and minimize intra-writer variation. While traditional Dynamic Time Warping (DTW)~\cite{kholmatov2005identity} excels at temporal alignment, its non-differentiable nature precludes direct end-to-end integration. To circumvent this, we employ soft-DTW~\cite{cuturi2017soft} during training to compute differentiable pairwise alignment costs within a triplet loss. During inference, standard DTW is utilized to calculate the optimal alignment and similarity scores between query and reference signatures for authentication.

\begin{figure}[t]
\includegraphics[width=\textwidth]{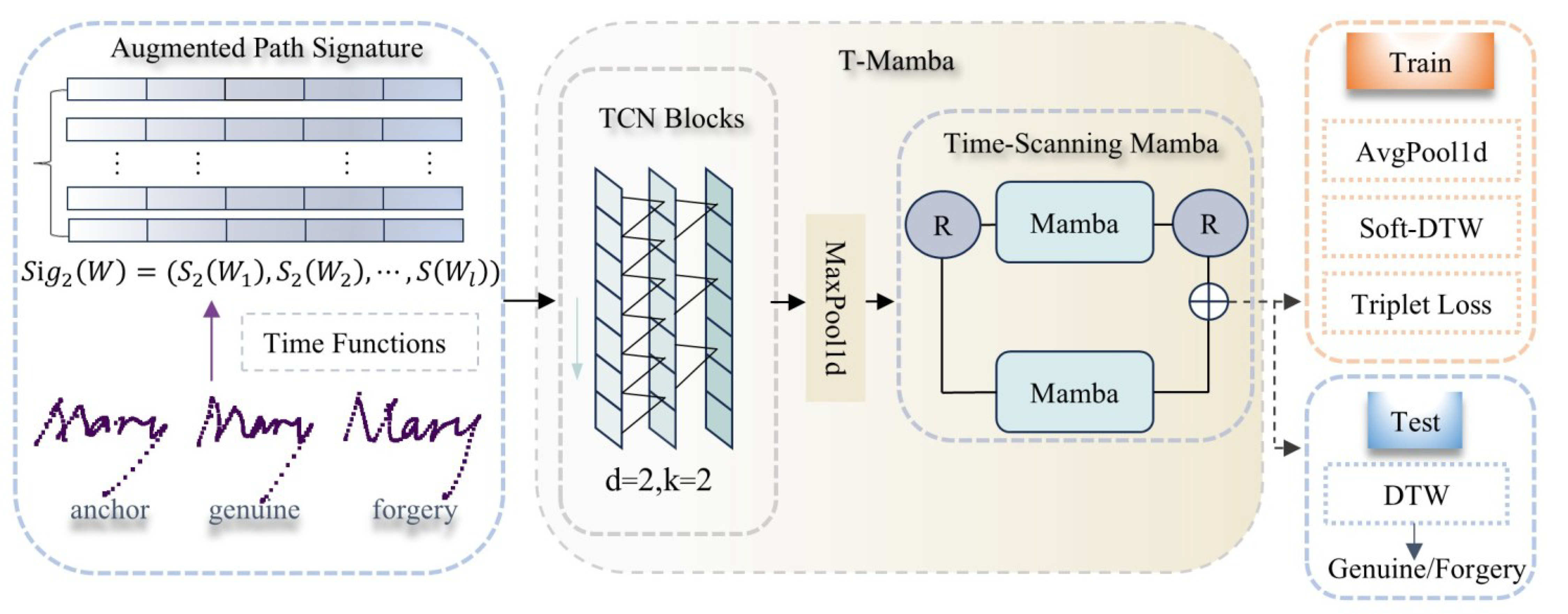}
\caption{\small{Overall framework of the proposed method.}} \label{fig1}
\end{figure}

The overall framework of our proposed method is illustrated in Fig.~\ref{fig1}.  Specifically, we first extract the APS features to obtain representations of the local geometric structure. These features are then fed into T-Mamba, optimised during training with a soft-DTW triplet loss, and evaluated during testing using standard DTW. The primary contributions of this paper are as follows:
\begin{itemize}
  \item We introduce the APS descriptor, a robust representation designed to capture the local geometric structure of signatures. By integrating path augmentations with sliding-window path signatures from rough path theory, it enables the extraction of high-order discriminative features.
  \item To the best of our knowledge, this work is the first to apply a state space model, specifically Mamba, to online signature verification and to demonstrate its effectiveness.
  \item We propose T-Mamba, a novel backbone that integrates TCN blocks with a time-scanning Mamba through max pooling. This design strikes an effective balance between local patterns and global contextual information.

  \item We achieve state-of-the-art performance on three public benchmarks (MCYT-100, SVC-2004 Task 2, DeepSignDB), demonstrating the effectiveness and robustness of our framework across diverse datasets and acquisition devices.
\end{itemize}
\vspace{-5mm}

\section{Related Work}
Over the past three decades, signature verification has remained an active research area, with comprehensive surveys~\cite{plamondon2000online,impedovo2012handwritten} reviewing its developments. In this pipeline, preprocessing and robust feature representation are essential to mitigate noise and handle variability. For instance, Li et al.~\cite{li2019stroke} proposed a stroke segmentation approach to isolate individual writing behaviors, while Lai et al.~\cite{lai2017online} introduced the LNPS descriptor, achieving scale and rotation invariance to significantly enhance feature robustness.
The path signature was originally introduced by Chen~\cite{chen1957integration} to study the geometry of paths, and later further developed by Lyons in the rough path theory~\cite{lyons1998differential}.
Graham introduced path signature features to machine learning~\cite{graham2013sparse} and won the ICDAR 2013 Online Isolated Chinese Character Recognition Competition.

DTW is a cornerstone for aligning signature sequences by minimizing cumulative alignment costs. To improve precision, SM-DTW~\cite{parziale2019sm} assigns higher weights to stable regions while penalizing distortions. Alternatively, to mitigate template sensitivity, Okawa~\cite{okawa2020singletemplate} introduced a single-template DTW approach that computes an Euclidean barycenter to summarize multiple genuine signatures into a representative reference. Furthermore, the development of differentiable  soft-DTW \cite{cuturi2017soft,jiang2022dsdtw} has successfully bridged the gap between DTW and deep learning. By incorporating soft-DTW into the loss function, Jiang et al.~\cite{jiang2022dsdtw}  directly optimised the alignment process during training, leading to significant performance gains.

To model the temporal dynamics,
Lai and Jin~\cite{lai2019recurrent} proposed a recurrent adaption network as a neural filter to learn discriminative representations from signature sequences.
To further exploit local features, researchers have proposed models based on CNNs~\cite{bhowal2022two,vorugunti2019online}, achieving lower EERs than earlier sequence models.
In~\cite{vorugunti2019online}, Vorugunti et al. introduced a lightweight framework employing depth-wise separable convolutions which effectively reduces parameters while maintaining high accuracy.
To better capture the global context, Transformer-based architectures have attracted growing attention. However, despite their dominance in natural language processing, their efficacy in OSV remains limited~\cite{gautam2023tsosvnet,melzi2023exploring}.
TSOSVNet~\cite{gautam2023tsosvnet}  reported a relatively high EER of 3.07\% on MCYT-100.  Melzi et al.~\cite{melzi2023exploring} demonstrated that a vanilla Transformer encoder can marginally surpass RNN-based baselines, while the added architectural complexity often results in a performance plateau that lags behind state-of-the-art models.

To overcome these limitations,
state-space models (SSMs)~\cite{gu2021combining,gu2024mamba} have emerged as a promising alternative, offering linear computational complexity while maintaining  long-range dependency modelling capabilities.
In 2024, Gu and Dao~\cite{gu2024mamba} introduced Mamba, a selective SSM that filters noise and retains task-relevant information in long sequences. Its superior balance between performance and complexity has established it as a potent sequence modelling backbone~\cite{ahamed2025tscmamba}. Building upon this, this paper presents the first attempt to apply Mamba to signature verification, achieving state-of-the-art results.

\section{Preprocessing with Augmented Path Signature}
\begin{figure}[H]
\centering
\includegraphics[width=0.7\textwidth]{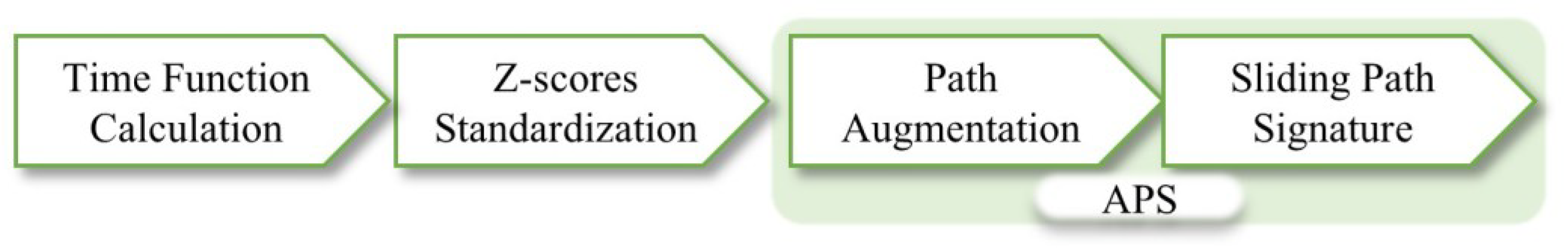}
\caption{Data preprocessing workflow.}
\label{fig2}
\end{figure}

Our preprocessing workflow is illustrated in Fig.~\ref{fig2}.  APS is a geometric descriptor constructed by first applying time and basepoint augmentations, then extracting sliding-window path signatures.
\subsection{Time Functions}
We select 12 time functions derived from coordinates and pressure, as shown in Table~\ref{timefunctions}, and then apply z-score normalization to each feature. Previous studies~\cite{jiang2022dsdtw,martinez2014mobile} have employed these time functions, demonstrating strong empirical performance for both stylus- and finger-written signatures.
\begin{table}[th!]
\caption{Time functions for dynamic feature extraction.}\label{timefunctions}
\centering
\begingroup
\setlength{\tabcolsep}{8pt}
\renewcommand{\arraystretch}{1.1}
\setlength{\heavyrulewidth}{0.6pt}   
\setlength{\lightrulewidth}{1.0pt}   
\begin{tabular}{ll}
\toprule
Number &Time Functions \\
\midrule
$1\sim3$ & Velocity features: $v_x, v_y, v = \sqrt{v_x^2 + v_y^2}$ \\
$4\sim6$ & Path orientation features: $\theta = \arctan(v_y / v_x), \cos(\theta), \sin(\theta)$ \\
$7\sim8$ & First-order derivatives of $v$ and $\theta$: $\dot{v}, \dot{\theta}$ \\
$9\sim10$ & Log curvature radius and centripetal acceleration: $\rho = \log(v / \dot{\theta}), c = v \cdot \dot{\theta}$ \\
$11\sim12$ & Total acceleration and pressure: $a = \sqrt{\dot{v}^2 + c^2}, p$ \\
\bottomrule
\end{tabular}
\endgroup
\end{table}
\subsection{Augmented Path Signature (APS)}
\subsubsection{Path Augmentations}
Since path signature features are invariant to translations and time reparametrizations, these properties may sometimes reduce their discriminative ability. Here we add two path augmentations to improve sensitivity to timestamps and the absolute spatial position respectively~\cite{chevyrev2026primer}.

The time augmentation transforms a $d$-dimensional path $X$ into a $(d+1)$-dimensional path by incorporating the increasing timestamps:
\begin{equation*}
\label{eq:time_augmentation}
\phi_t(X) = \big((t_0, X_0), (t_1, X_1), \ldots, (t_l, X_l)\big).
\end{equation*}
This augmentation encodes the writing speed, and can ensure the uniqueness of the path signature~\cite{hambly2010uniqueness}.

The basepoint augmentation adds a zero at the beginning of the series:
\begin{equation*}
\phi_b(X)=\big(0,X_1,\ldots,X_l\big).
\end{equation*}
It encodes the absolute position and removes the translation invariance introduced by the path signature features.

The effect of these augmentations, both individually and in combination, will be discussed  in Section~\ref{experiments}.

\subsubsection{Sliding-Window Path Signatures}
Let $X: [a, b] \to \mathbb{R}^d$ be a continuous path of bounded variation. For any positive integer $N$, the order-$N$ truncated path signature is defined as:
\begin{equation*}
S_N(X) = \Big(\{S(X)^{(j_1)}\}_{j_1 =1}^{d}, \{S(X)^{(j_1,j_2)}\}_{j_1,j_2 = 1}^{d}, \ldots, \{S(X)^{(j_1,\ldots,j_N)}\}_{j_1,\ldots,j_N =1}^{d}\Big),
\end{equation*}
where
\begin{equation*}
S(X)^{(j_1, \ldots, j_k)} = \int_{a \le t_1 < \cdots < t_k \le b}
 \mathrm{d}X^{j_1}_{t_1} \cdots \mathrm{d}X^{j_k}_{t_k},
\end{equation*}
which is a feature map of dimension $\sum_{k=1}^{N}d^k$. In the experiments, the continuous path is constructed by linearly interpolating between consecutive time points.

\begin{figure}[H]
\centering
\includegraphics[width=0.5\textwidth]{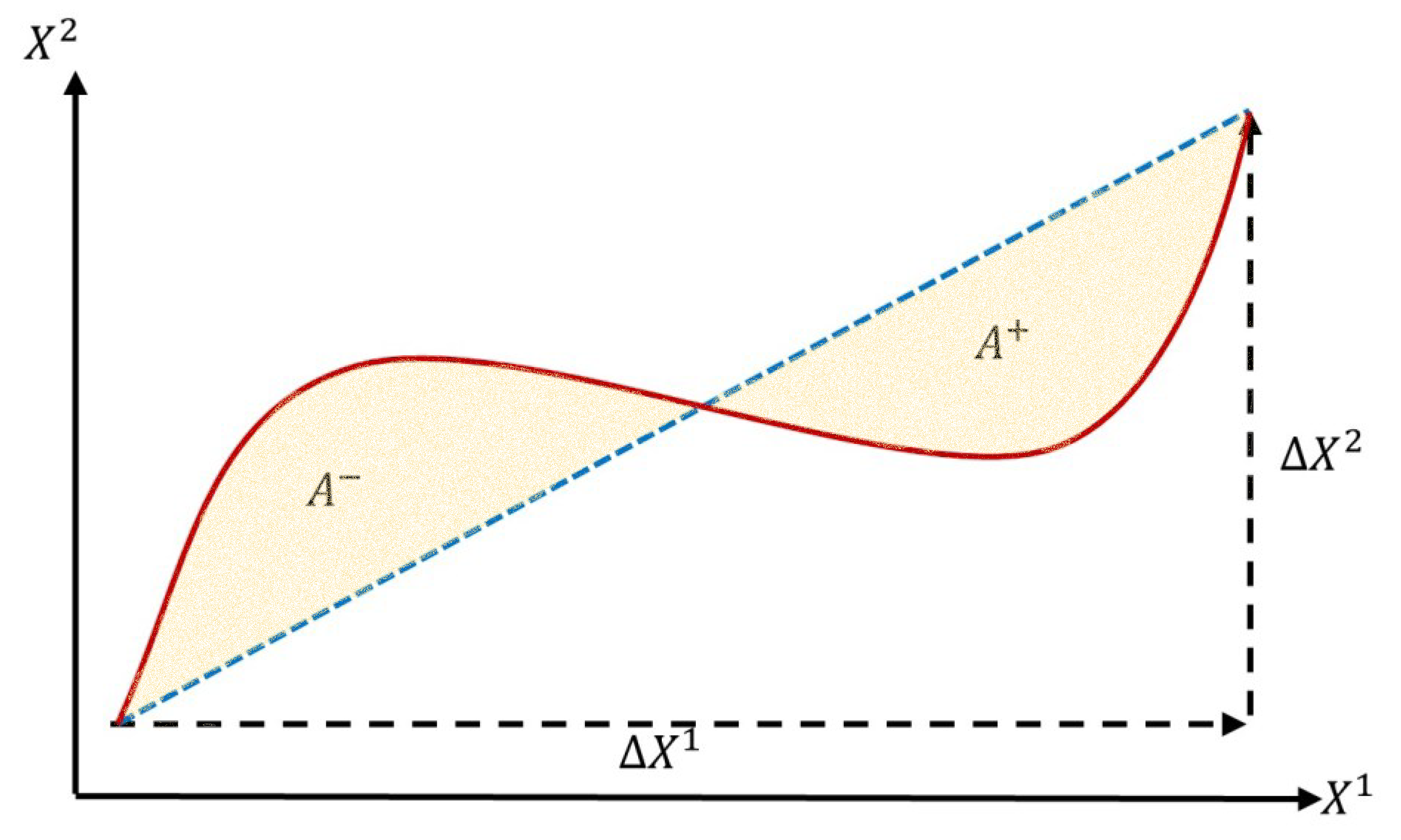}
\caption{Geometric interpretation of the order-2 truncated path signature of a planar path (shown in red) that starts at the bottom-left and ends at the top-right.}
\label{fig3}
\end{figure}

An illustration of the order-2 truncated path signature of a planar path is shown in Fig.~\ref{fig3}. The oriented increments $\Delta X^1$ and $\Delta X^2$  correspond to the first-level iterated integrals, while the signed area $A^{+}-A^{-}$ is a linear combination of second-level iterated integrals based on Green's theorem.

\subsubsection{Benefits of the Path Signature }
\begin{itemize}
\item The path signature collapses the temporal dimension into a fixed-dimensional representation, enabling direct comparison of varying-length time series.
\item It offers a top-down geometric description, with the truncation improving computational efficiency and suppressing noisy local variations.
\item Iterated integrals encode nonlinear inter-channel interactions, thereby capturing fine-grained dynamic features, which are not accessible to linear transformations.
\end{itemize}

To encode contextual information, a sliding window of size $w$ is applied to the time series at each time step $t$ before extracting path signature features.
Define the sliding window $W_t$:
\begin{equation*}
W_t = (X_t, X_{t+1}, \dots, X_{t+w-1}).
\end{equation*}
The  APS descriptor of order $N$ is then defined as the sequence of path signatures computed over each sliding window of the augmented path:
\begin{equation*}
\text{APS}:= \big( S_{N}(W_1),S_{N}(W_2),S_{N}(W_3), \ldots, S_{N}(W_l) \big),
\end{equation*}
where $l$ is the total number of windows.

\section{T-Mamba with Soft-DTW}
\subsection{TCN Block}
\begin{figure}
\centering
\subfloat[TCN block\label{figtcn}]{%
    \includegraphics[height=6cm]{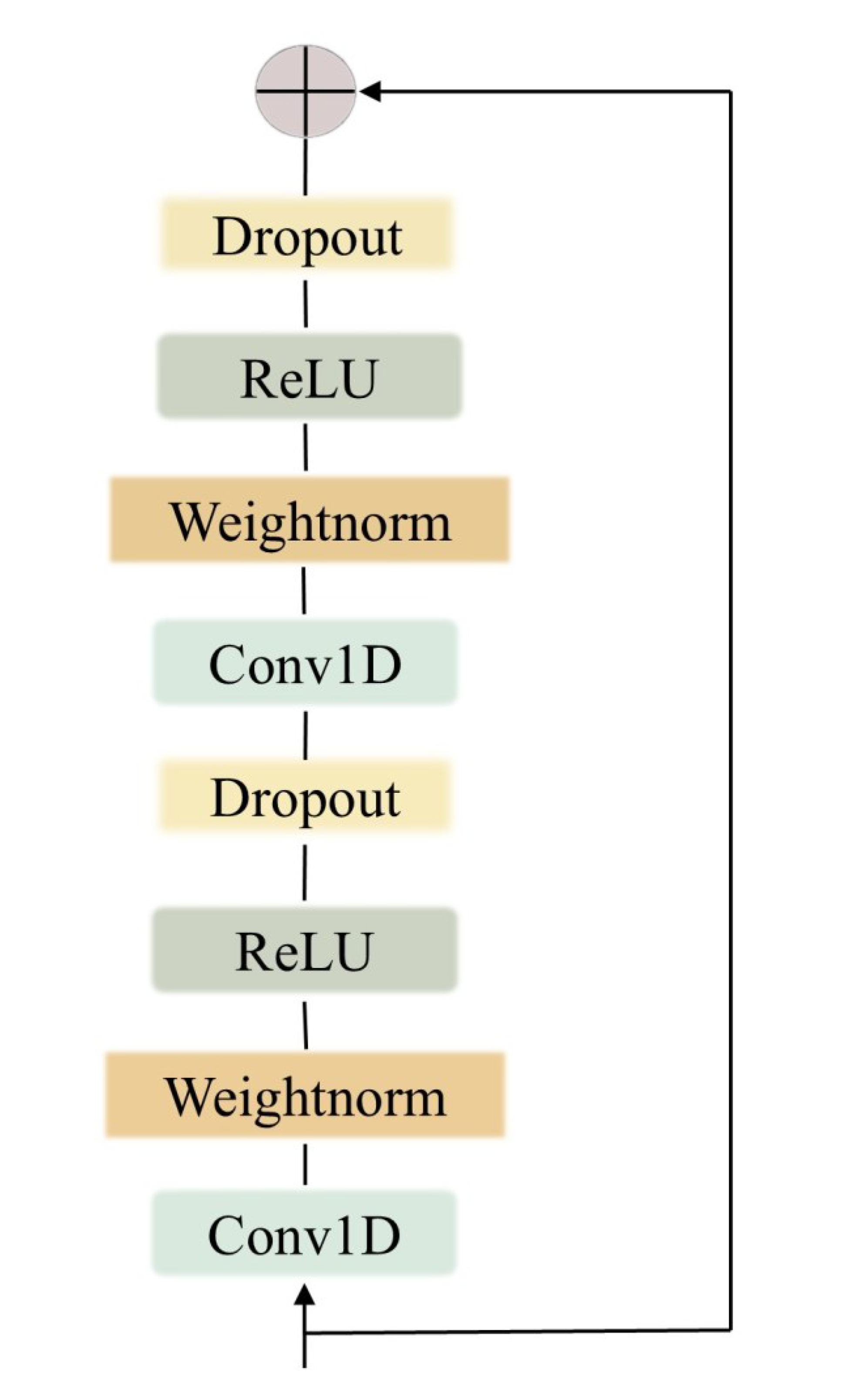}
}
\hspace{0.08\textwidth}
\subfloat[Mamba\label{figmamba}]{%
    \includegraphics[height=6cm]{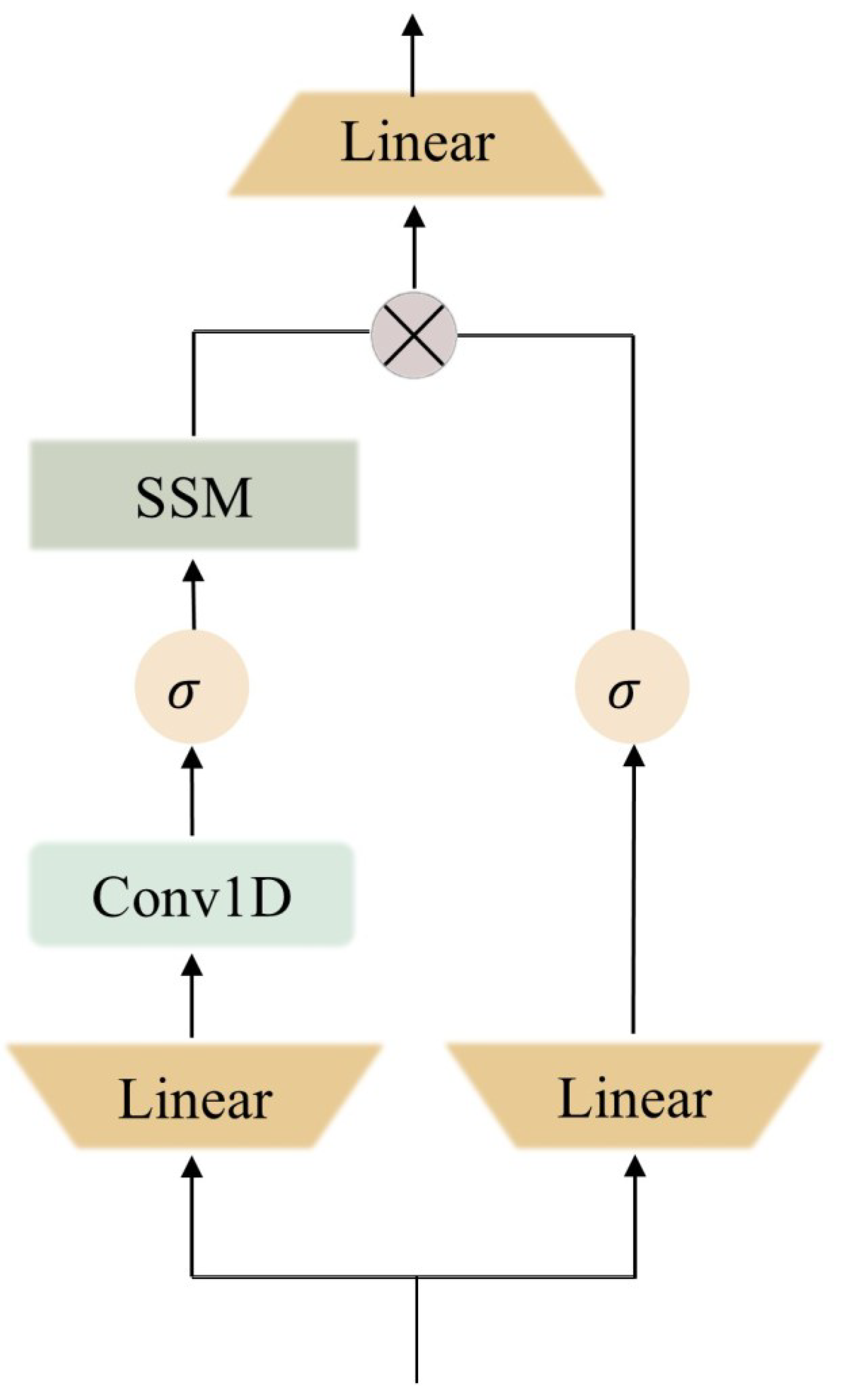}
}
\caption{The structure of (a) TCN block and (b) Mamba. The symbol $\oplus$ indicates an element-wise addition and $\otimes$ represents matrix multiplication. Here $\sigma$ is SiLU.}
\label{figmamtcn}
\end{figure}

The temporal convolutional network (TCN)~\cite{BaiTCN2018} is depicted in Fig.~\ref{figtcn}. This block is a residual module composed of two layers of dilated causal 1D convolutions. Each Conv1D layer is followed by weight normalization, a ReLU activation, and a spatial dropout for regularization.
The receptive field of the network is jointly determined by the network depth $m=2$, filter size $k=2$ and dilation factor $d=2$, where each filter tap contributes $(k-1)d$ steps of historical context that accumulate across stacked layers.
\vspace{-5mm}
\subsection{Time-Scanning Mamba}\label{mamba}
The structure of Mamba is illustrated in Fig.~\ref{figmamba}.
Continuous state space models define the evolution of a latent state $h(t)$ and output $y(t)$ as:
\begin{equation*}
  \label{eq:lti-ssm}
  \frac{dh(t)}{dt} = A h(t) + B x(t),\qquad y(t) = C h(t),
\end{equation*}
Here $A$, $B$, $C$ are learnable and time-invariant, and the system is inherently linear time-invariant (LTI)~\cite{gu2021combining}.
 To execute the continuous models in practice, they are mapped to a discrete recurrence $h_k = \tilde A h_{k-1} + \tilde B x_k$ through a fixed discretization rule $\Delta$, where the exact forms of  $\tilde A$ and $\tilde B$ depend on the numerical scheme used.

Mamba's key innovation is to remove the LTI constraint by making the SSM parameters $B$, $C$, and the discretization rule $\Delta$ input-dependent and time-varying, introducing selectivity into the model.
For an input token $x_k$, these parameters are computed through linear maps and activation functions:
\begin{equation*}\label{bt}
  B_k = W_D x_k, \quad C_k = W_D x_k, \quad \Delta_k = \operatorname{softplus}\left(\text{bias} + W_D(W_1 x_k)\right),
\end{equation*}
where $W_d$ is a linear map to a $d$-dimensional space and the bias is learnable.
Specifically, based on~\cite[Theorem 1]{gu2024mamba}, when $D=1$, $A=-1$, $B=1$, the selective recurrence takes the form:
\begin{equation*}\label{gate}
  g_k=\operatorname{sigmoid}\!\big(\mathrm{Linear}(x_k)\big),\qquad
h_k=(1-g_k)\,h_{k-1}+g_k\,x_k.
\end{equation*}
This gating mechanism enables precise state control: as $g_k \to 1$, the update focuses on the current input $x_k$, effectively resetting the state. To circumvent the convolutional constraints imposed by this selective recurrence, Mamba leverages a hardware-aware parallel algorithm to maintain high throughput~\cite{gu2024mamba}.

However, since this selective recurrence is strictly causal, Mamba cannot exploit future context, which limits its efficacy when the prior context is uninformative. Inspired by the reversal-based scanning in~\cite{ahamed2025tscmamba}, we propose a time-scanning Mamba that captures complementary bidirectional contexts via a parameter-shared dual temporal scan.

Given a sequence $a = [a_1, \ldots, a_L] \in \mathbb{R}^{L \times D}$, its temporal reversal is denoted as $\mathrm{R}(a) = [a_{L}, a_{L-1}, \ldots, a_1]$. Let $\mathrm{Mamba}_\theta(a)$ represent the output of a vanilla Mamba parameterized by $\theta$. To capture bidirectional temporal context using parameter-shared scanning, the sequence is processed in both directions and aggregated via element-wise addition $\oplus$:
\begin{equation*}\label{time-scanning}
\begin{aligned}
    \mathrm{output}^{(1)} &= a \;\oplus\; \mathrm{Mamba}_\theta(a), \\
    \mathrm{output}^{(2)} &= \mathrm{R}(a) \;\oplus\; \mathrm{Mamba}_\theta(\mathrm{R}(a)), \\
    \mathrm{output} &= \mathrm{output}^{(1)} \;\oplus\; \mathrm{R}(\mathrm{output}^{(2)}).
\end{aligned}
\end{equation*}
Here, the first two branches execute forward and backward scanning with shared weights $\theta$, maintaining original information via residual paths. Aggregating these streams allows each token to leverage both past and future contexts for enhanced sequence modelling.
\subsection{Soft-DTW with a Triplet Loss}
\subsubsection{$\gamma$-Soft-DTW Alignment Cost}
Let $X=[x_1,\ldots,x_{l_1}]\in\mathbb{R}^{{l_1}\times d}$ and $Y=[y_1,\ldots,y_{l_2}]\in\mathbb{R}^{{l_2}\times d}$ be two variable-length sequences that are produced by T-Mamba. Let $\mathcal{B}_{\ell_1,\ell_2}\subset\{0,1\}^{\ell_1\times \ell_2}$  denote the set of admissible binary alignment matrices satisfying continuity, monotonicity and boundary conditions~\cite{sakoe1978dynamic}. The DTW alignment cost is defined as
\begin{equation}\label{dtw}
  d(X,Y)=\operatorname{DTW}(X,Y)
  \;=\;
  \min_{B\in\mathcal{B}_{\ell_1,\ell_2}}
  \big\langle B,\Delta(X,Y)\big\rangle,
\end{equation}
where $\Delta(X,Y)$ is an $\ell_1\times\ell_2$ matrix given by $[\Delta(X,Y)]_{i,j}:=\Vert x_i-y_j\Vert^2_2$ and $\langle\cdot,\cdot\rangle$ denotes the inner product.

However, DTW cannot be directly optimized via backpropagation
due to its non-differentiable nature. To address this limitation,  Cuturi and Blondel~\cite{cuturi2017soft} introduced a differentiable alternative based on a smoothed minimum controlled by a non-negative parameter $\gamma$:
\begin{equation*}
  \min\nolimits_{\gamma}\{b_1,\ldots,b_n\}=
  \begin{cases}
    \min_i b_i, & \gamma=0,\\
    -\gamma\log\!\sum_{i=1}^n e^{-b_i/\gamma}, & \gamma>0,
  \end{cases}
\end{equation*}
and  the $\gamma$-soft-DTW alignment cost is defined as:
\begin{equation*}
  d_\gamma(X,Y) = \operatorname{DTW}_\gamma(X,Y)
  \;=\;
  \min\nolimits_{\gamma}\Big\{
    \big\langle B,\Delta(X,Y)\big\rangle:\;
    B\in\mathcal{B}_{\ell_1,\ell_2}
  \Big\}.
\end{equation*}
The parameter $\gamma$ controls the smoothness of the cost. As $\gamma \to 0$,  it recovers the original DTW. For $\gamma > 0$,  the cost becomes differentiable, allowing gradients to propagate across all warping paths.
\subsubsection{Triplet Loss with Alignment Cost}
We formulate OSV as an alignment-based learning task, employing a triplet loss to enforce intra-writer similarity and maximize inter-writer separability. For each training triplet $(X_a, X_g, X_f) \in \mathcal{S}$ comprising an anchor, a genuine sample, and a forgery—the objective minimizes the internal alignment costs. Specifically, to enhance intra-writer compactness, we incorporate an intra-writer contraction term that penalizes the average alignment cost over the anchor-genuine subset $\mathcal{S}_g$, defined as:
\begin{equation*}
  \mathcal{V}_{\mathrm{intra}}
  =
  \frac{1}{|\mathcal{S}_g|}\sum_{(X_a,X_g)\in\mathcal{S}_g} d_\gamma(X_a,X_g).
\end{equation*}
Then  the final loss function is given by:
\begin{equation}\label{eqxi}
  \mathcal{L}
  \;=\;
  \frac{1}{|\mathcal{S}|}\sum_{(X_a,X_g,X_f)\in\mathcal{S}}
  \Big(\Big[d_\gamma(X_a,X_g)+\xi-d_\gamma(X_a,X_f)\Big]_+
  \;+\;
  \lambda \mathcal{V}_{\mathrm{intra}}\Big),
\end{equation}
where the first term enforces a positive margin $\xi$ between the anchor–genuine and anchor–forgery alignment costs. Concurrently, the intra-writer term, regulated by $\lambda$, minimizes variance among genuine samples to regularize the embedding space. This dual constraint yields compact writer-specific clusters and establishes a robust discriminative boundary.
\subsubsection{Verifier Based on DTW}During the test stage, given $m$ reference signatures ${X^1,\ldots,X^m}$ and a test signature $Y$, we compute the DTW cost $d(X_i,Y)$ for $i=1,\cdots,m$ based on~Eq.~\ref{dtw}. Let $D$ denote the average DTW cost over all pairs of reference signatures. For each writer, we compute the following scores:
\begin{equation*}
  s_{\mathrm{ave}}(Y)=\frac{1}{m}\sum_{i=1}^{m} d(X_i,Y)/ \sqrt{D},
  \quad
  s_{\min}(Y)=\min_{i=1,\ldots,m} d(X_i,Y)/ \sqrt{D}.
\end{equation*}
Define the final score $s(Y)=s_{\mathrm{ave}}(Y)+s_{\min}(Y)$. If $s(Y)\le \tau$ where $\tau$ is a writer-dependent threshold, the test signature $Y$ is classified as a genuine; otherwise, it is considered a forgery. The performance of the OSV system is evaluated using the writer-specific equal error rate (EER), a standard criterion for OSV, obtained by sweeping $\tau$ to the point where the false acceptance rate equals the false rejection rate.
\section{Experiments}\label{experiments}
\subsection{Datasets and Implementation Details}
\vspace{-0.8mm}
Experiments were conducted on three benchmark datasets: MCYT-100~\cite{ortega2003mcyt} (100 writers, 25 genuine and 25 forged signatures per writer), SVC-2004 Task 2~\cite{yeung2004svc2004} (40 writers, 20 genuine and 20 forged signatures per writer), and the large-scale DeepSignDB~\cite{tolosana2021deepsign} (69,972 signatures from 1,526 writers). DeepSignDB integrates multiple subsets (e.g., MCYT, BiosecurID, Biosecure DS2, e-BioSign DS1, and e-BioSign DS2) to provide diverse stylus- and finger-written conditions.

Following~\cite{vorugunti2020osvfusenet}, we evaluate the system under both skilled forgery (S\shortus N) and random forgery (R\shortus N) protocols, where $N \in \{5, 10, 15\}$ denotes the number of genuine training samples per writer. For enrollment, 5 genuine signatures serve as references, with the remaining samples reserved for verification.

The T-Mamba framework comprises two TCN blocks (hidden dimensions: 256 and 128) followed by a time-scanning Mamba module (state expansion: 256). All models were implemented in PyTorch and trained on a single NVIDIA A100 GPU (80~GB) using SGD for 20 epochs. The initial learning rate was set to 0.001 with an exponential decay factor of 0.9 per epoch, utilizing a batch size of 40. Following~\cite{jiang2022dsdtw}, the margin factor $\xi$ in Eq.~\ref{eqxi} and the smoothing parameter $\gamma$ were fixed at 1 and 5, respectively. The configuration was empirically optimized via grid search, with all results averaged over five random seeds.

\subsection{Effectiveness of APS}
To evaluate the contribution of the APS descriptor, we conducted a comparative experiment on MCYT-100 under S\shortus 05. We assessed the performance of APS across various sliding-window sizes $w$ using four augmentation configurations: (\romannumeral1) the path signature (PS), (\romannumeral2)  the time augmentation (TA) then PS, (\romannumeral3) the basepoint augmentation (BA) then PS, and (\romannumeral4) APS.

Table~\ref{tab1:signature} shows that a moderate window size ($w=9$ to $13$) achieves optimal performance by encoding a balanced temporal context that captures sufficient local dynamics while avoiding excessive smoothing or fragmentation. Beyond window scale, the results confirm the efficacy of the proposed APS descriptor, where both time and basepoint augmentations outperform vanilla PS in most cases. Although APS does not always deliver the best performance across all window sizes, it attains the lowest overall EER at the optimal window size of $w = 11$, suggesting a strong synergistic effect at this particular scale.
\begin{table}[H]
\caption{Performance comparison across different augmentation configurations and sliding-window sizes $w$.}
\label{tab1:signature}
\centering
\begingroup
\setlength{\tabcolsep}{6pt}
\renewcommand{\arraystretch}{1.0}
\setlength{\heavyrulewidth}{0.6pt}   
\setlength{\lightrulewidth}{1.0pt}   
\begin{tabular}{lcccccc}
\toprule
EER\% & \textbf{$w=5$} & \textbf{$w=7$} & \textbf{$w=9$} & \textbf{$w=11$} & \textbf{$w=13$} & \textbf{$w=15$} \\
\midrule
PS        & 0.700 & 0.688 & 0.656 & 0.638 & 0.604 & 0.662 \\
TA+PS      & 0.696 & 0.676 & 0.628 & 0.631 & 0.619 & 0.616 \\
BA+PS & 0.705 & 0.673 & 0.608 & 0.635 & 0.606 & 0.640 \\
\textbf{APS}    & 0.677 & 0.661 & 0.626 & \textbf{0.555} & 0.625 & 0.625 \\
\bottomrule
\end{tabular}
\endgroup
\end{table}

\begin{figure}[!htbp]
\centering
\includegraphics[height=5cm]{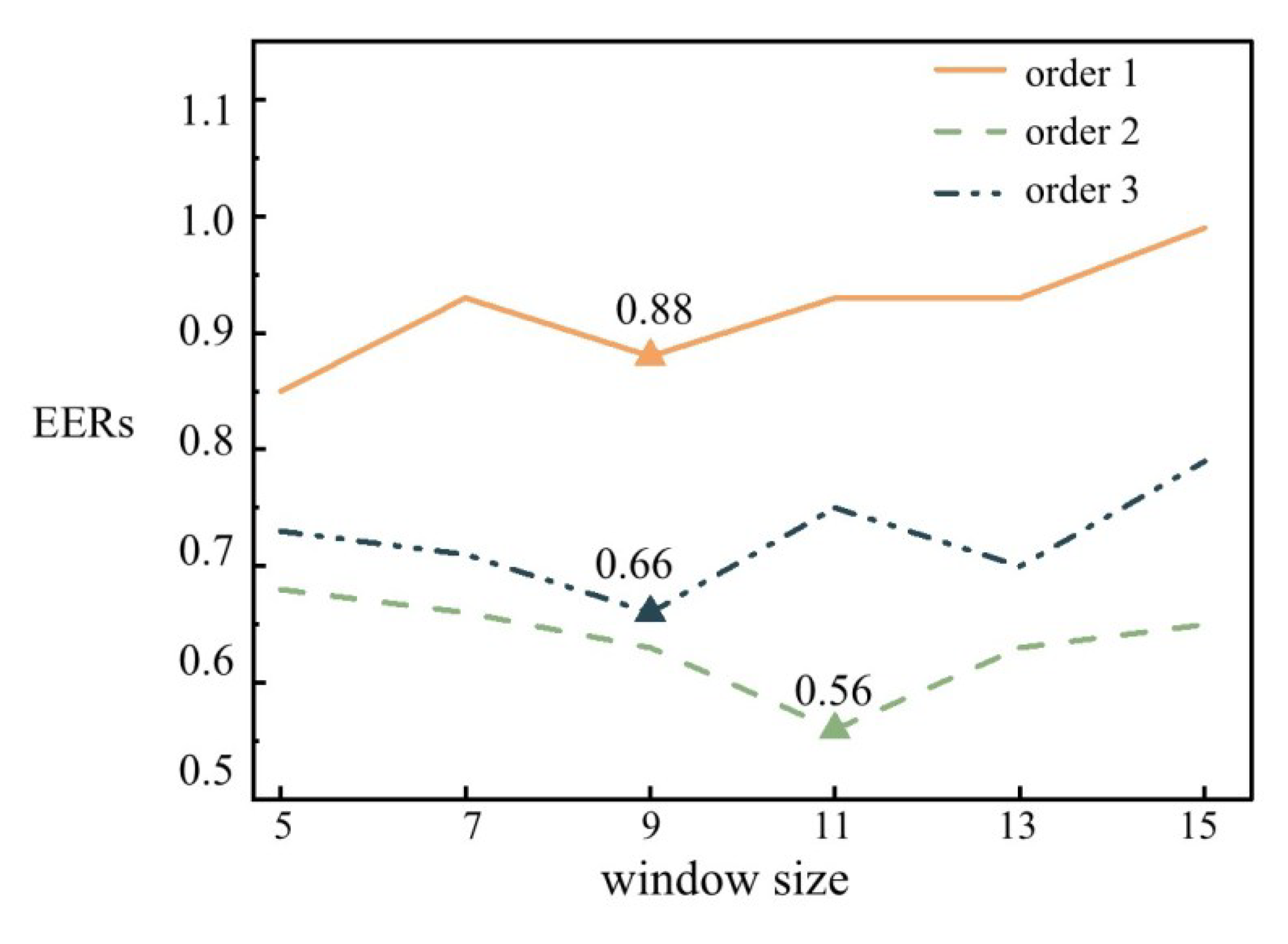}
\caption{Performance comparison across different orders of the path signature in the APS descriptor.}
\label{sigorder}
\end{figure}

We also compared APS performance under different truncation orders of the path signature, as illustrated in Fig.~\ref{sigorder}. The results show that the second-order path signature achieves the best performance across all window sizes, followed by the third-order, while the first-order performs the worst.
This trend confirms that higher order terms can capture inter-channel interactions that are crucial for effective signature verification. However, increasing the order beyond $N=2$ adds unnecessary noisy details that degrade performance.
\vspace{-4mm}
\subsection{Evaluation of the Proposed Method}
To evaluate computational efficiency, we compared the proposed time-scanning Mamba with the standard Mamba and Transformer architectures on MCYT-100 under S\shortus05, as shown in Table~\ref{tab:efficiency_compact}. Compared with the Transformer, the Mamba variants reduce the number of parameters, FLOPs, and runtime GPU memory consumption by more than $60\%$, while maintaining comparable training speed. In addition, the proposed time-scanning mechanism reduces the EER from $0.87\%$ to $0.56\%$, demonstrating its enhanced modelling capability.

\begin{table}[!htbp]
\centering
\small
\newcolumntype{A}{>{\hsize=1.0\hsize\centering\arraybackslash}X} 
\newcolumntype{B}{>{\hsize=1.4\hsize\centering\arraybackslash}X} 
\newcolumntype{C}{>{\hsize=0.9\hsize\centering\arraybackslash}X} 
\newcolumntype{D}{>{\hsize=0.7\hsize\centering\arraybackslash}X} 

\caption{Comparison of computational efficiency and verification performance on MCYT-100 under S\shortus05.}\label{tab:efficiency_compact}
\begin{tabularx}{\columnwidth}{lAABCD}
\toprule
Model & Parameters & FLOPs & GPU Memory & Time per & EER \\
      & (M)        & (G)   & Usage (MB) & Epoch (s)          & (\%) \\
\midrule
Time-scanning Mamba
  & 0.33 & 0.29 & 3,902 & 21.6 & 0.56 \\
Mamba  & 0.33 & 0.29 & 3,452 & 18.5 & 0.87\\
Transformer & 0.83          & 0.73          & 11,250         & 19.0 & 0.83 \\
\bottomrule
\end{tabularx}
\end{table}

We conducted a cross-ablation study to validate the effectiveness of the APS descriptor and the T-Mamba model on MCYT-100 under S\shortus 05, SVC-2004 Task 2 under S\shortus05, and DeepSignDB. The path signature truncation order is 2 and the sliding-window size is 11. We adopt DsDTW with CRAN~\cite{jiang2022dsdtw} as a strong baseline, as it achieved first place in the ICDAR 2021 Competition on On-Line Signature Verification~\cite{Tolosana2021ICDARSVC}. Table \ref{tab:ablation of t-mamba} shows that T-Mamba consistently outperforms CRAN on all datasets, with or without APS. Incorporating APS further improves performance and stabilises results for both models, as reflected in the lower standard deviations. These findings confirm that the APS descriptor provides robust feature representations, and T-Mamba serves as a strong discriminative backbone for OSV. The performance of CRAN on DeepSignDB suggests that APS may be sensitive to the noise inherent in finger-written signatures, which CRAN does not sufficiently suppress. By contrast, T-Mamba’s selective state modelling exploits path signature features more effectively, resulting in improved performance.
\begingroup
\small
\setlength{\tabcolsep}{6.2pt}
\begin{table}[!htbp]
\caption{ Ablation study on the effect of APS and a comparison between T-Mamba and CRAN (EER\%). `w'
and `w/o' denote `with' and `without'.}\label{tab:ablation of t-mamba}
\centering
\begingroup
\small
\begin{tabular}{lcccccc}
\toprule
\multirow{2}{*}{Method} & \multirow{2}{*}{ APS} & \multirow{2}{*}{MCYT-100} & \multirow{2}{*}{SVC-2004 Task 2} & \multicolumn{2}{c}{DeepSignDB} \\
\cmidrule(lr){5-6}
 &  &  &  & Stylus & Finger \\
\midrule
\multirow{2}{*}{T-Mamba} & w   & $\mathbf{0.56\pm0.04}$  & $\mathbf{1.39\pm0.23}$& $\mathbf{0.76\pm0.05}$ & $\mathbf{3.20\pm0.15}$ \\
                         & w/o & $0.93\pm0.08$  & $2.30\pm0.25$ & $0.92\pm0.08$ & $3.67\pm0.18$ \\
\cmidrule(lr){1-6}
\multirow{2}{*}{CRAN}    & w   & $0.67\pm0.06$  & $2.25\pm0.35$ & $0.88\pm0.08$ & $4.23\pm0.32$ \\
                         & w/o & $1.15\pm0.07$ & $3.11\pm0.42$ & $1.00\pm0.11$ & $3.46\pm0.39$ \\
\bottomrule
\end{tabular}
\endgroup
\end{table}
\endgroup

We compared our framework against other state-of-the-art methods on MCYT-100 across skilled (S\shortus05, S\shortus10, S\shortus15) and random (R\shortus05, R\shortus10) forgery configurations. As summarized in Table~\ref{tab:mcyt_eer}, our method achieves state-of-the-art results across all skilled forgery protocols, yielding EERs of 0.56\%, 0.45\%, and 0.43\%, respectively. This performance scales robustly with training sample volume, indicating enhanced generalization. For random forgeries, the framework attains competitive EERs of 0.04\% (R\shortus05) and 0.03\% (R\shortus10). While Probabilistic-DTW~\cite{al2019quantifying} reports 0.01\% on R\shortus05, its lack of evaluation on other protocols limits comprehensive comparison. In particular, Few-shot learning~\cite{vorugunti2019online} performs poorly under skilled forgeries despite reasonable random-forgery EERs, suggesting that skilled forgery is the key challenge on MCYT-100 and the proposed method addresses it effectively.

\begin{table}[t]
\caption{Comparison between the proposed method and other state-of-the-art methods on MCYT-100 (EER\%).}
\label{tab:mcyt_eer}
\centering
\begingroup
\setlength{\tabcolsep}{4pt}
\renewcommand{\arraystretch}{0.98}
\setlength{\heavyrulewidth}{0.6pt}   
\setlength{\lightrulewidth}{1.0pt}   
\small
\begin{tabularx}{\linewidth}{@{}lcccccc@{}}
\toprule
Method & S\shortus05 & S\shortus10 & S\shortus15 & R\shortus05 & R\shortus10\\
\midrule
\textbf{Proposed Method} & \textbf{0.56} & \textbf{0.45} & \textbf{0.43} & 0.04 & \textbf{0.03}  \\
Time-series averaging + DTW~\cite{okawa2021time} & 0.72 & -- & -- & 0.07 & --\\
SynSig2Vec~\cite{lai2020synsig2vec} & 0.93 & -- & -- & -- & --  \\
DTW and warping path-based features~\cite{sharma2017exploration} & 1.15 & 1.08 & 0.84 & 0.13 & --  \\
Mean templates and multiple DTW distances~\cite{okawa2020singletemplate} & 1.28 & -- & -- & -- & -- \\
Enhanced contextual DTW~\cite{sharma2016enhanced} & 1.55 & -- & -- & -- & -- \\
DsDTW with CRAN~\cite{jiang2022dsdtw} & 1.01 & -- & -- & -- & -- \\
Modified DTW with signature curve constraint~\cite{xia2018discriminative} & 2.17 & -- & -- & -- & -- \\
Interval valued symbolic~\cite{guru2017interval} & 2.20 & -- & -- & 1.00 & --  \\
TSOSVNet~\cite{gautam2023tsosvnet} & 3.07 & 1.23 & 0.45 & -- & --  \\
SM-DTW using distance normalization~\cite{parziale2019sm} & 3.09 & 2.25 & -- & 1.30 & --  \\
Feature fusion based on deep learning~\cite{vorugunti2020osvfusenet} & 3.02 & 1.83 & 1.25 & 0.42 & 0.10 \\
DTW and sigma-lognormal analysis~\cite{fischer2016signature} & 3.56 & -- & -- & 1.01 & --\\
Few-shot learning~\cite{vorugunti2019online} & 7.03 & 5.70 & 3.95 & 0.05 & 0.06  \\
Probabilistic-DTW~\cite{al2019quantifying} & -- & -- & -- & \textbf{0.01} & --  \\
\bottomrule
\end{tabularx}
\endgroup
\end{table}

Similar experiments were conducted on SVC-2004 Task 2 under S\shortus05, S\shortus10, R\shortus05, R\shortus10. As shown in Table~\ref{tab:svc2004_eer}, the proposed method delivers the best or near-best performance across the four settings. It achieves the lowest EER on S\shortus10 and attains perfect separation in both random-forgery cases. On the more challenging skilled-forgery setting S\shortus05 with fewer references, the proposed method ranks the second, surpassed only by Few-shot learning~\cite{vorugunti2019online}, which is specifically designed for low-shot training.  Notably, increasing the number of genuine training samples from 5 to 10 leads to a substantial improvement in the skilled scenario (1.39\% to 0.25\%), indicating that the model effectively leverages additional training data.

\begin{table}[!htp]
\caption{Comparison between the proposed method and other state-of-the-art methods on SVC-2004 Task 2 (EER\%).}
\label{tab:svc2004_eer}
\centering
\begingroup
\setlength{\tabcolsep}{6pt}
\renewcommand{\arraystretch}{0.98}
\setlength{\heavyrulewidth}{0.6pt}   
\setlength{\lightrulewidth}{1.0pt}   
\begin{tabular}{lcccc}
\toprule
Method & S\shortus05 & S\shortus10 & R\shortus05 & R\shortus10 \\
\midrule
\textbf{Proposed Method} & 1.39 & \textbf{0.25} & \textbf{0.00} & \textbf{0.00} \\
Few-shot learning~\cite{vorugunti2019online} & \textbf{0.87} & 0.35 & 1.40 & 0.15 \\
DsDTW with CRAN~\cite{jiang2022dsdtw} & 1.62 &-- & -- & --  \\
TSOSVNet~\cite{gautam2023tsosvnet} & 1.70 & 0.75 & -- & -- \\
Time-series averaging + DTW~\cite{okawa2021time} & 2.08 & 1.53 & 0.11 & 0.03 \\
DTW and warping path-based features~\cite{sharma2017exploration} & 2.53 & 2.79 & -- & 0.00 \\
Mean templates and multiple DTW distances~\cite{okawa2020singletemplate} & 2.98 & 1.80 & -- & -- \\
Enhanced contextual DTW~\cite{sharma2016enhanced} & 2.73 & -- & -- & -- \\
Modified DTW with signature curve constraint~\cite{xia2018discriminative} & 2.60 & -- & -- & -- \\
SynSig2Vec~\cite{lai2020synsig2vec} & 2.63 & -- & -- & -- \\
Feature fusion based on deep learning~\cite{vorugunti2020osvfusenet} & 3.93 & 2.98 & 0.45 & 0.39 \\
Stroke point warping~\cite{kar2017stroke} & -- & 1.00 & -- & -- \\
Probabilistic-DTW~\cite{al2019quantifying} & -- & -- & 0.00 & -- \\
\bottomrule
\end{tabular}
\endgroup
\raggedright
\end{table}

\subsection{Experiments with DeepSignDB}
To further assess the effectiveness and generalization capability of the proposed method, we conducted experiments on DeepSignDB, currently the largest public dataset for OSV, following the standard protocol~\cite{tolosana2021deepsign}.
We adopt DsDTW~\cite{jiang2022dsdtw} as a strong baseline, which won first place in the ICDAR 2021 Competition on On-Line Signature Verification~\cite{Tolosana2021ICDARSVC}. The EER\textsubscript{global} is computed based on a uniform threshold for all writers in the subset. The overall EER\textsubscript{global} is computed using a single global threshold shared by all writers within the Stylus and Finger subsets, respectively. In Table~\ref{tab:deepsign}, the EER\textsubscript{global}s for DsDTW are taken directly from~\cite{jiang2022dsdtw}.
The writer-specific EERs for DsDTW are not reported  in~\cite{jiang2022dsdtw} and were obtained by running their publicly available code.
Here we tune the window size $w$ on each subset for optimal performance.

As shown in Table~\ref{tab:deepsign}, the proposed method either outperforms or remains comparable to DsDTW on most acquisition devices, achieving lower overall EER\textsubscript{global}s and EERs. Notably, in the Stylus setting, our method attains an EER of $0.76\%$, yielding a $24\%$ relative error reduction over DsDTW. Despite minor fluctuations in a few configurations, the aggregate gains across diverse subsets demonstrate strong cross-dataset and cross-device generalization against pronounced device variability. These improvements are primarily driven by the synergy between APS, which captures robust local geometric structures, and T-Mamba, which models bidirectional long-range dependencies. Furthermore, unlike the sequential, recurrence-based CRAN in DsDTW~\cite{jiang2022dsdtw}, T-Mamba leverages a hardware-friendly parallel execution mechanism, significantly enhancing backbone efficiency.

\begin{table}[!htbp]
\caption{Comparison of DsDTW and the proposed method on DeepSignDB in terms of EER\textsubscript{global}\% and EER\%.}\label{tab:deepsign}
\centering
\begingroup
\small
\setlength{\tabcolsep}{3pt}
\renewcommand{\arraystretch}{0.98}
\setlength{\heavyrulewidth}{0.6pt}   
\setlength{\lightrulewidth}{1.0pt}   
\setlength{\cmidrulewidth}{0.6pt}    
\begin{tabularx}{\linewidth}{@{}%
  l@{\hspace{4pt}}%
  l@{\hspace{5pt}}%
  l@{\hspace{1pt}}%
  c@{\hspace{6pt}}%
  Z@{\hspace{6pt}}%
  Z@{\hspace{6pt}}%
  Z@{\hspace{6pt}}%
  Z@{}}
\toprule
Type & Dataset & Device & $w$ &
\multicolumn{2}{c}{EER\textsubscript{global}} &
\multicolumn{2}{c}{EER} \\
\cmidrule(r){5-6}\cmidrule(l){7-8}
 &  &  &  & DsDTW & Ours & DsDTW & Ours \\
\midrule
\multirow{10}{*}{Stylus}
 & MCYT         & Wacom Intuos A6            & 11 & 1.86 & 1.99 & 1.01& 0.80 \\
 & BiosecurID   & Wacom Intuos 3             & 13 & 0.95 & 0.98 & 0.33 & 0.21 \\
 & Biosecure DS2& Wacom Intuos 3             &  5 & 2.66 & 2.89 & 1.59 & 1.50 \\
 & e-BioSign DS2& Wacom STU-530              & 11 & 0.71 & 0.71 & 0.25 & 0.19 \\
 & \multirow{5}{*}{e-BioSign DS1}
                 & Wacom STU-500             & 10 & 3.54 & 1.79 & 0.14 & 0.71 \\
 &               & Wacom STU-530             & 10 & 3.24 & 1.67 & 0.90 & 0.00\\
 &               & Wacom DTU-1031            & 10 & 4.44 & 3.67 & 1.33 & 0.90 \\
 &               & Samsung ATIV 7            & 10 & 4.14 & 2.19 & 2.20 & 0.38 \\
 &               & Samsung Galaxy Note 10.1  & 10 & 5.08 & 4.67 & 1.11 & 1.38 \\
 & Overall EER &                     &    & 2.54 & \textbf{2.44} & 1.00 & \textbf{0.76} \\
\midrule
\multirow{5}{*}{Finger}
 & \multirow{2}{*}{e-BioSign DS1}
                 & Samsung ATIV 7            &  17  & 11.45 & 11.32   & 7.45 & 7.19   \\
 &               & Samsung Galaxy Note 10.1  & 17   & 9.57  & 6.89   & 5.11 & 4.71   \\
 & \multirow{2}{*}{e-BioSign DS2}
                 & Samsung Galaxy Note 10.1  & 5   & 2.53  & 0.36  & 0.92 & 0.25   \\
 &               & Samsung Galaxy S3         &  5  & 4.29  & 3.76   & 0.36 & 0.64   \\
 & Overall EER&                     &    & 6.99  & \textbf{6.10}   & 3.46 & \textbf{3.20}   \\
\bottomrule
\end{tabularx}
\endgroup
\end{table}

\section{Conclusion}
We developed a novel OSV framework that integrates the augmented path signature (APS) descriptor with the T-Mamba model. Experimental results demonstrate that, with an appropriate window size and truncation order, APS captures complex inter-channel interactions and produces discriminative features. Furthermore, the T-Mamba model serves as a new backbone for OSV by integrating TCN blocks with a time-scanning Mamba to model local and global dependencies and leverage past and future contexts. Overall, our framework achieves state-of-the-art performance on MCYT-100, SVC-2004 Task 2, and the largest dynamic signature database to date DeepSignDB. The strong performance observed on smaller datasets (e.g. e-BioSign) warrants further investigation, particularly in the context of few-shot learning.

\vspace{-1.0em}
\subsection*{Acknowledgements}
DY gratefully acknowledges the support of the National Natural Science Foundation of China (Young Scholar Grant 12201081), Chongqing University (Starting Grant 02080011044104, School of Mathematics and Statistics Basic Discipline Development Fund 0208005406001) and the Engineering and Physical Sciences Research Council (Programme Grant EP/S026347/1).

%
%
%
\bibliographystyle{splncs04}
\bibliography{mybibliography}
\end{document}